\documentclass[10pt,twocolumn,letterpaper]{article}

\usepackage[pagenumbers]{cvpr} %

\usepackage{graphicx}
\usepackage{amsmath}
\usepackage{amssymb}
\usepackage{booktabs}
\usepackage{xcolor}         %
\usepackage{listings}
\usepackage{makecell}
\usepackage{tablefootnote}
\usepackage{float}

\usepackage[pagebackref,breaklinks,colorlinks]{hyperref}

\usepackage[capitalize]{cleveref}
\crefname{section}{Sec.}{Secs.}
\Crefname{section}{Section}{Sections}
\Crefname{table}{Table}{Tables}
\crefname{table}{Tab.}{Tabs.}

\def\cvprPaperID{*****} %
\def\confName{CVPR}
\def\confYear{2023}

\newcommand{\ourmethod}{Palm}

\DeclareMathOperator*{\argmax}{arg\,max}

\usepackage{pifont} 
\newcommand{\cmark}{\ding{51}}%
\newcommand{\xmark}{\ding{55}}%

\begin{document}

\title{\ourmethod: Predicting Actions through Language Models \\
@ Ego4D Long-Term Action Anticipation Challenge 2023}

\author{
Daoji Huang \hspace{10mm}
Otmar Hilliges \hspace{10mm}
Luc Van Gool  \hspace{10mm}
Xi Wang \\
ETH Z\"urich 
}

\maketitle

\begin{abstract}
We present \ourmethod, a solution to the Long-Term Action Anticipation (LTA) task utilizing vision-language and large language models. Given an input video with annotated action periods, the LTA task aims to predict possible future actions. We hypothesize that an optimal solution should capture the interdependency between past and future actions, and be able to infer future actions based on the structure and dependency encoded in the past actions. Large language models %
have demonstrated remarkable commonsense-based reasoning ability. Inspired by that, \ourmethod~chains an image captioning model and a large language model. It predicts future actions based on frame descriptions and action labels extracted from the input videos. Our method outperforms other participants in the EGO4D LTA challenge and achieves the best performance in terms of action prediction. 
Our code is available at \url{https://github.com/DanDoge/Palm}.
\end{abstract}

\section{Introduction}
\label{sec:intro}

Predicting future actions from egocentric videos is inherently a challenging task given the uncertainty of the future, and very often there exist multiple plausible action candidates and executive orders. Previous solutions~\cite{Ego4D, vclip} rely on Transformer-based feature extraction followed by multiple classification heads which predict verbs and nouns independently. 
ICVAE\cite{ICVAE} proposes to use a predicted scenario, which summarizes the activity depicted in the video, as a condition for action prediction, but such data-driven models are inherently limited by the dataset and data distribution it is trained on. 
We argue that a shared feature is not enough to predict future actions as it cannot model the complex dependency of verbs and nouns within the same action and the dependency across past and future actions. 
We hypothesize that leveraging commonsense knowledge embedded in large language models can help us discover the underlying structure and dependency of different activities and make better predictions of future actions.  

We propose \ourmethod~to predict actions through large language models (LLMs). 
We first extract essential information from the input video through an image captioning module and an action recognition module. 
The captions and recognized actions are then used to create a prompt which is fed into a large language model for future action anticipation. 
The variability of the predictions is achieved by sampling multiple solutions from LLMs. 
We present our approach in detail in Section~\ref{sec:method}, followed by experimental analysis in Section~\ref{sec:exp}.

\begin{figure}[t]
  \centering
  \includegraphics[width=\linewidth]{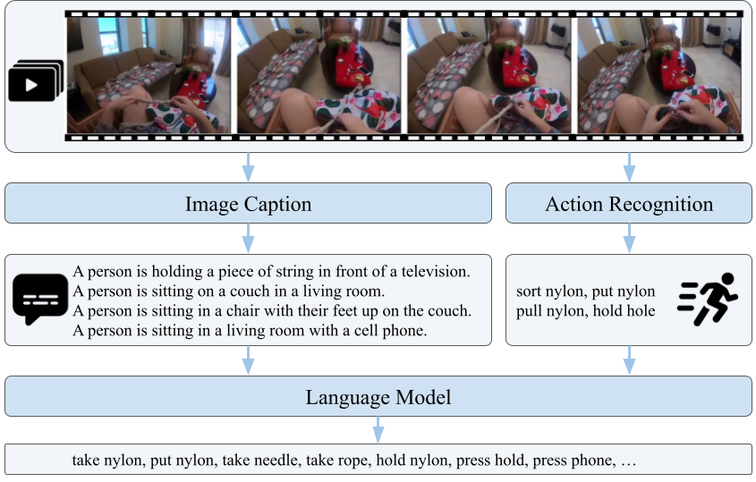}
  \caption{\textbf{\ourmethod:} Predicting Actions through Language Models. Given an input video, \ourmethod~uses an image captioner and an action recognizer to describe its content. The obtained image captions and action labels are combined to create a prompt, which is inputted into a large language model. A sequence of future actions is then predicted by the large language model.  
  }
  \label{fig:teaser}
\end{figure}

\section{Method}
\label{sec:method}
Our idea is to use natural language to describe past events and perform reasoning and prediction in the semantic space, leveraging the commonsense knowledge embedded in large language models. 
Figure~\ref{fig:teaser} shows an overview of the proposed method \ourmethod.

\subsection{Task Definition}
The task of long-term action anticipation (LTA) involves predicting a sequence of $20$ future actions in chronological order from an input video clip that spans approximately 5 minutes.  
In addition, the segment boundaries of each action in the videos are also provided as part of the input.  
Each action is described by a (verb, noun) pair.

\subsection{Prompt Design}

We formulate the action anticipation task as a sentence completion task and predict future actions by prompting large language models (LLMs) with past action descriptions. 
Out-of-the-box large language models do not work well on the LTA task in a zero-shot setting. 
Following the common practice, we design the prompt with a few examples to maximize the reference given to the model.
Figure~\ref{fig:prompt} shows the template of our prompts, which consists of an instruction, a few examples, followed by a query.  
A prefix instruction is to adapt the model to the LTA task. 
Each example contains $N$ full-sentence narrations and corresponding $N$ action labels of (verb, noun) pairs. 
We also give the future 20 actions in the examples. 
Similarly, a query is composed of narrations and action labels of the past $N^\prime$ actions. 
We use a larger number of action observations $N^\prime > N $ in the query. 

\begin{figure}[htb]
\centering
\begin{tabular}{l}

\begin{lstlisting}
You are going to complete an action 
sequence, an action is one (verb, noun) 
pair. A complete sequence consists of 
28 actions. You will also be given a 
text description of the past actions 
for reference. 

Narrations: [caption] x N 
Actions: [action] x (N+Z)

...

Narrations: [caption] x N 
Actions: [action] x (N+Z)

Narrations: [caption] x N' 
Actions: [action] x N'
\end{lstlisting}
 \end{tabular}
\caption{\textbf{Prompt template} we used to query large language models. The prompt consists of an instruction paragraph, a few examples, and a query.}
\label{fig:prompt}
\end{figure}

\paragraph{Past actions.} 
Accurate action descriptions are crucial as they implicitly provide the corresponding context of the video input to large language models and give possible action candidates.  
We adopt the Transformer model used in the Ego4D~\cite{Ego4D} baseline to recognize the actions from the observed video clips.
We use video features extracted from EgoVLP~\cite{EgoVLP}, a vision-language model trained on the Ego4D dataset and widely used in various Ego4D challenges. 
A Transformer is used to extract a single feature vector from the input video features, followed by two classification heads to predict the verbs and nouns.

\paragraph{Narrations.} The past action labels do not capture all the visual information and context, e.g. the location of the actions or the background objects. 
Our idea is to use an image captioning model to capture the action context, providing more information related to the visual input. %
To do so, we use the given action segments and generate a caption from each past action. 
More specifically, we use the middle frame of each action segment and generate a caption starting with the prefix ''\textit{A person is}".

\paragraph{Prompt selection.}
Inspired by a maximal-marginal-relevance-based (MMR) exemplar selection strategy~\cite{mmr}, we aim to select a set of examples that are semantically relevant to the query while being diverse enough to provide nonrepetitive information. 
Specifically, we iteratively select a set of examples $p_i \in T$ from the training set $D$ that are semantically close to the query prompt $q$ but also diverse enough to provide additional information.

\begin{equation}
\label{eq:mmreq}
p_j = {\argmax}_{p_j\in D/T} \lambda \mathcal{S} (q, p_j) - (1-\lambda) \max_{p_i \in T} \mathcal{S}(p_i, p_j),   
\end{equation}
where $S$ measures the semantic similarity between two descriptions and the parameter $\lambda$ is to balance between similarity and diversity. We use MPNet~\cite{mpnet} to extract textural embeddings and cosine similarity to measure semantic similarity.

\subsection{Language Model Inference}

Large language models will output comma-separated verb-noun pairs following the format in the prompt. 
We then extract the verb-noun pairs and append them to our prediction if both verb and noun fall in the Ego4D label space. For predictions that have less than 20 actions, we pad it with the last action.
We sample LLMs multiple times to obtain $K$ predictions.

\section{Experiments}
\label{sec:exp}

We train the action recognition model on version 2 of the Ego4D dataset and use the EgoVLP~\cite{EgoVLP} features of four past actions as input (dim=3072). 
Training configurations are otherwise the same as in the Ego4D baseline. 
We use GPT-Neo-1.3B~\cite{gpt-neo} as the large language model and BLIP2-opt-2.7B~\cite{BLIP2} as the image captioning model. 
The prompt includes 8 examples with $N = 8$ past actions and $Z = 20$ future actions. 
The query has $N' = \min(12, \mbox{action\_idx})$ previous actions, where $\mbox{action\_idx}$ is the index of last observed action. 
In other words, $\mbox{action\_idx}$ corresponds to the total number of actions contained in the input video. 
Following the requirement of Ego4D, we predict an action sequence starting from the action with $\mbox{action\_idx} = 7$ and generate five sequences for each video.

\paragraph{Evaluation metric.}

We evaluate our method using the metric of edit distance. 
Given $K$ predicted sequences with each consisting $Z$ future actions $\{\{(\hat n_{z,k}^{}, \hat v_{z,k}^{})\}_{z=1}^Z\}_{k=1}^K$, the action edit distance is measured as the best individual edit distance with the ground truth action sequence $\{(n_z^{}, v_z^{})\}_{z=1}^Z)$, 

\begin{equation}
    \Delta_{E_{tot}}
 =\min\limits_{k=1..K} \Delta_{E}(\{(\hat n_{z,k}^{}, \hat v_{z,k}^{})\}_{z=1}^Z, \{(n_z^{},  v_z^{})\}_{z=1}^Z).
\end{equation}
Similar metrics can be defined for verb and noun predictions respectively.

\subsection{Leaderboard results}

\begin{figure*}
\hsize=\textwidth
\centering
\includegraphics[width=0.9\linewidth]{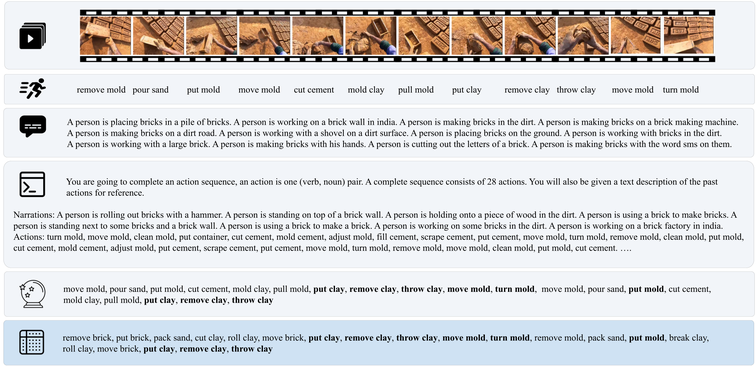}
\caption{\textbf{Qualitative result} of our framework with the input images, the prompt and one prediction. Ground truth labels are shown in the last row.}
\label{fig:visual}
\end{figure*}

Tabel~\ref{table:leaderboard} shows the comparison of our method to the Ego4D baseline on the test set.
We achieve first place in the leaderboard of the CVPR 2023 competitions. Figure~\ref{fig:visual} shows one example. Notice the similarity between the query captions and the retrieved examples and how the output captures the variability of future actions.

\begin{table}
  \caption{Results on the Ego4D test set.}
  \label{table:leaderboard}
  \centering
  \small
  \begin{tabular}{lccc}
    \toprule
    Method     & Verb ED $\downarrow$     & Noun ED $\downarrow$ & Action ED $\downarrow$\\
    \midrule
    Ego4D baseline & 0.7169 & 0.7359 & 0.9253 \\
    Brunos & 0.7004 & 0.7092 & 0.9142 \\
    PaMsEgoAI & \textbf{0.6838} & 0.6785 & 0.8933 \\
    \midrule
    \ourmethod & 0.6956 & \textbf{0.6506} & \textbf{0.8856} \\
    \bottomrule
  \end{tabular}
\end{table}

\subsection{Ablation}

We examine the effectiveness of our main contributions in this section. 
Here the version 2 validation set is used. 
We compare two different caption models vit-gpt2~\cite{vitgpt2} and BLIP2~\cite{BLIP2} in Table~\ref{table:ablation:caption}.
For overall actions, BLIP2 yields better results, especially in terms of noun prediction. 
Further adding MMR prompt selection yields the best result over all metrics.

\begin{table}
  \caption{Ablation of captioning models.}
  \label{table:ablation:caption}
  \centering
  \small
  \begin{tabular}{llccc}
    \toprule
    Model & \thead{Prompt \\ Selection} & Verb ED   $\downarrow$  & Noun ED $\downarrow$ & Action ED $\downarrow$\\
    \midrule
    vit-gpt2~\cite{vitgpt2} & random     & 0.7168 & 0.7026 & 0.9046 \\
    BLIP2~\cite{BLIP2}    & random     & 0.7235 & 0.6908 & 0.9025 \\
    BLIP2~\cite{BLIP2}    & MMR        & \textbf{0.7165} & \textbf{0.6767} & \textbf{0.8934} \\
    \bottomrule
  \end{tabular}
\end{table}

Table~\ref{table:ablation:prompt} shows the importance of using narrations and actions in the prompts. While both parts are beneficial, past actions seem to be more important in providing object-related context, resulting in lower edit distance in terms of noun and action predictions.

\begin{table}
  \caption{Results of ablating the prompt content.}
  \label{table:ablation:prompt}
  \centering
  \small
  \begin{tabular}{ccccc}
    \toprule
    Narrations & Actions & Verb ED  $\downarrow$    & Noun ED  $\downarrow$& Action ED  $\downarrow$\\
    \midrule
    \xmark    & \cmark    & 0.7275 & 0.6798 & 0.9006 \\ %
    \cmark     & \xmark   & \textbf{0.7152} & 0.7338 & 0.9206 \\ %
    \cmark     & \cmark & 0.7165 & \textbf{0.6767} & \textbf{0.8934} \\
    \bottomrule
  \end{tabular}
\end{table}

Table~\ref{table:ablation:llm} shows the benefit of using a larger language model which allows for a larger prompt size. We compare GPT-Neo-1.3B~\cite{gpt-neo} with all variants in the GPT-2~\cite{gpt2} family where the number of examples is maximized within the token limit. The increasing model size consistently lowers verb and noun errors and thus has better action edit distance. With more model parameters and larger prompt size, GPT-Neo-1.3B outperforms all GPT-2 variants. An even larger language model would likely lead to further improvements.

\begin{table}
  \caption{Results of ablating the language models.}
  \label{table:ablation:llm}
  \centering
  \small
  \setlength\tabcolsep{2.7pt} %
  \begin{tabular}{lcccc}
    \toprule
    Model & \# Example & Verb ED $\downarrow$      & Noun ED $\downarrow$  & Action ED $\downarrow$  \\
    \midrule
    GPT-2-small  &3    & 0.7507 & 0.6851 & 0.9090 \\
    GPT-2-medium  &3   & 0.7369 & 0.6906 & 0.9074 \\
    GPT-2-large    &3  & 0.7228 & 0.6814 & 0.9027 \\
    GPT-2-xl       &3  & 0.7296 & 0.6795 & 0.8983 \\
    GPT-Neo-1.3B   &8 & \textbf{0.7165} & \textbf{0.6767} & \textbf{0.8934} \\
    \bottomrule
  \end{tabular}
\end{table}

\subsection{Limitations}

\begin{figure}[t]
  \centering
  \includegraphics[width=\linewidth]{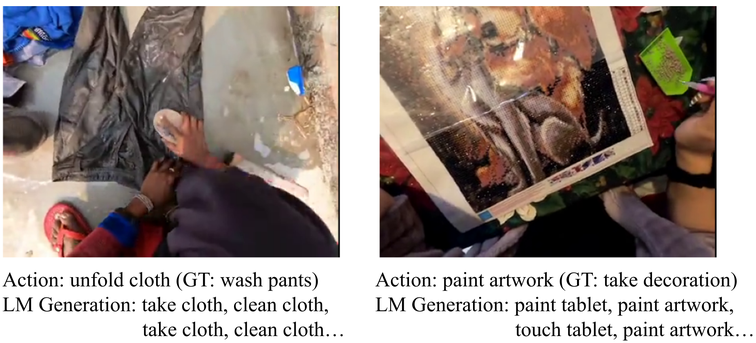}
  \caption{
    \textbf{Failure case examples} when action recognition fails.
  }
  \label{fig:limit}
\end{figure}

Our model is limited by the quality of recognized past actions. 
More specifically, edit distance gets larger when the accuracy of recognized actions is lower. Table~\ref{table:limit} shows a strong negative correlation between the accuracy of past action recognition and edit distance. We measure the accuracy as the mean accuracy of the actions, i.e. the percentage of recognized actions being equal to ground truth actions. Figure~\ref{fig:limit} shows typical failure cases where a wrongly recognized noun leads to incorrect prediction, resulting in edit-distance being 1, i.e. all predicted actions are incorrect.

\begin{table}
  \caption{Linear regression coefficient of action recognition accuracy and edit distance. The standard deviations of coefficients are shown in brackets.}
  \label{table:limit}
  \centering
  \small
  \begin{tabular}{ccc}
    \toprule
     & Verb Accuracy   & Noun Accuracy\\
    \midrule
    Verb ED      & -0.1298 (0.003) &  0.0160 (0.002) \\
    Noun ED      & -0.1012 (0.005) & -0.1835 (0.004) \\
    Action ED    & -0.1057 (0.002) & -0.0725 (0.002) \\
    \bottomrule
  \end{tabular}
\end{table}

\section{Conclusion}

We present \ourmethod, a framework that leverages pretrained vision-language and large language models for long-term action anticipation. We demonstrate the effectiveness of performing the LTA task in the language domain and the benefits of using large vision-language and language models. However, a pre-trained large language model might not be able to predict future action sequences when past action descriptions are incorrect.

{\small
\bibliographystyle{ieee_fullname}
\bibliography{egbib}
}

\end{document}